\documentclass{article}

\usepackage{microtype}
\usepackage{graphicx}
\usepackage{booktabs} 

\usepackage{hyperref}

\usepackage[accepted]{icml2025}

\usepackage{amsmath}
\usepackage{amssymb}
\usepackage{mathtools}
\usepackage{amsthm}
\usepackage{color}
\usepackage{colortbl}
\usepackage{multirow} 
\usepackage[capitalize,noabbrev]{cleveref}
 
\theoremstyle{plain}

\theoremstyle{definition}

\theoremstyle{remark}

\usepackage[textsize=tiny]{todonotes}

\usepackage{subcaption}
\newlength\savewidth
\newcommand{\tablestyle}[2]{\setlength{\tabcolsep}{#1}\renewcommand{\arraystretch}{#2}\centering\footnotesize}

\icmltitlerunning{}

\begin{document}

\twocolumn[
\icmltitle{Task-agnostic Prompt Compression with Context-aware Sentence\\Embedding and Reward-guided Task Descriptor}

\begin{icmlauthorlist}
\icmlauthor{Barys Liskavets}{alterra}
\icmlauthor{Shuvendu Roy}{queensu}
\icmlauthor{Maxim Ushakov}{alterra}
\icmlauthor{Mark Klibanov}{workday}
\icmlauthor{Ali Etemad}{queensu}
\icmlauthor{Shane Luke}{workday}
\end{icmlauthorlist}

\icmlaffiliation{alterra}{Alterra AI, Palo Alto, United States.}
\icmlaffiliation{queensu}{Queen’s University, Canada.}
\icmlaffiliation{workday}{Workday Inc}

\icmlcorrespondingauthor{Barys Liskavets}{liskovets.borets@gmail.com}
\icmlkeywords{Machine Learning, ICML}

\vskip 0.3in
]

\printAffiliationsAndNoticeArxiv{}  

\begin{abstract}
The rise of Large Language Models (LLMs) has led to significant interest in prompt compression, a technique aimed at reducing the length of input prompts while preserving critical information.
However, the prominent approaches in prompt compression often require explicit questions or handcrafted templates for compression, limiting their generalizability. We propose Task-agnostic Prompt Compression (TPC), a novel framework that generalizes compression across tasks and domains without requiring input questions or templates. TPC generates a context-relevant task description using a task descriptor trained on a curated dataset of context and query pairs, and fine-tuned via reinforcement learning with a reward function designed to capture the most relevant information. The task descriptor is then utilized to compute the relevance of each sentence in the prompt to generate the compressed prompt. We introduce 3 model sizes (Base, Large, and Huge), where the largest model outperforms the existing state-of-the-art methods on LongBench and ZeroSCROLLS benchmarks, and our smallest model performs comparable to the existing solutions while being considerably smaller.
\end{abstract}

\section{Introduction}
The emergence of Large Language Models (LLMs) has spurred extensive research into prompting techniques, including chain-of-thought reasoning \cite{wei2022chain}, in-context learning \cite{dong2022survey}, and retrieval-augmented generation \cite{lewis2020retrieval} to harness LLMs' generalization and reasoning abilities for various applications. In practice, effective prompting often requires detailed and lengthy inputs to generate high-quality responses in domain-specific tasks. However, lengthy prompts significantly increase inference time and costs. 
To address this challenge, a new research direction called \textit{prompt compression} \cite{jiangllmlingua,jiang2023longllmlingua,CPC} has emerged, which focuses on reducing prompt length while retaining the critical information needed to accurately answer user queries.

Early research on prompt compression predominantly focused on token-level techniques, which involve analyzing the importance of individual tokens in a prompt and removing less informative ones to generate the compressed prompt. However, token-level compression often leads to incoherent or fragmented sentences, hindering the overall performance \cite{CPC}. To address this, recent approaches, such as CPC \cite{CPC}, have shifted to sentence-level compression, which evaluates the relevance of entire sentences in relation to the input question. While this method represents a significant improvement over token-level approaches, its reliance on explicit questions or manually crafted templates for non-question-based tasks (e.g. summarization or code completion) restricts its broader applicability.

\begin{figure}
    \centering
    \includegraphics[width=0.85\linewidth]{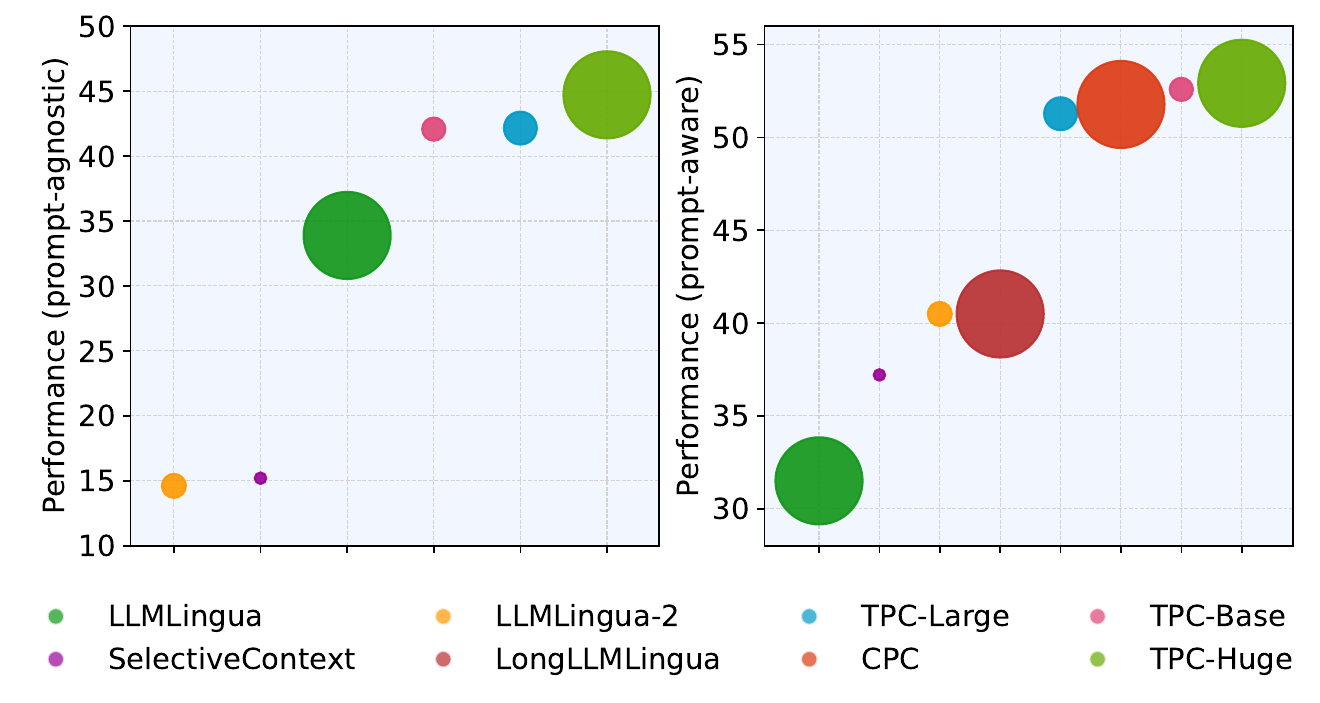}
    \caption{Comparison of model size versus performance for different prompt compression methods in both prompt-aware and prompt-agnostic setups. Our largest model, TPC-Huge, outperforms all existing methods while maintaining a comparable size to existing solutions. On the other hand, our smallest model, TPC-Base, achieves a competitive performance despite being significantly smaller in size. }
    \label{fig:bannar}
    \vspace{-10pt}
\end{figure}

In this work, we propose Task-agnostic Prompt Compression (TPC), a generic prompt compression method capable of compressing input prompts across tasks and domains without relying on input questions or handcrafted prompt templates. The central idea of TPC involves using a novel task descriptor to create a context-relevant task description, which facilitates compression by comparing its context-aware embedding similarity to each sentence in the prompt. Here, the task descriptor is a causal Language Model (LM), trained on our curated dataset of prompt-question pairs, that generates a context-relevant task description encapsulating the main concept of the prompt. Since the effectiveness of the prompt compression relies on the quality of the generated task description, we introduce a reinforcement learning (RL) approach to further fine-tune the task descriptor with a novel reward function designed to encourage capturing the most relevant information of the prompt in the task description. Finally, the context-aware embedding is computed using a context-aware sentence encoder trained on our curated dataset of multi-hop questions, answers, and positive and negative pairs.

We present comprehensive experiments to evaluate our proposed solution and compare it against existing methods on two evaluation setups: prompt-aware compression and prompt-agnostic compression on two popular benchmarks LongBench and ZeroSCROLLS adopted by existing literature on prompt compression. 
We report the results for three variants of our model size: TPC-Base, TPC-Large, and TPC-Huge, containing 0.5B, 1B, and 7B parameters, respectively. As summarized in Figure \ref{fig:bannar}, our TPC shows significant improvements over existing methods on both prompt-aware and prompt-agnostic setups. Our smallest model, while being considerably smaller in size, outperforms or performs comparable to the existing state-of-the-art (SOTA) methods. 
Overall, we make the following contribution in this paper:

\begin{enumerate}
    \item We introduce TPC, a task-invariant prompt compression method, which unlike existing methods doesn't require task-specific hand-crafted templates to achieve generalizable compression. TPC generates a context-relevant task description using a task descriptor LM that is used for compression by comparing the context-relevant embedding to each sentence in the prompt. 

    \item We propose a novel reward function and use RL-based fine-tuning of the task descriptor to capture the most relevant task descriptions.

    \item We curate two datasets required for training the context-relevant task descriptor and context-aware sentence encoder of our proposed TPC. 
    
    \item Our proposed solution outperforms existing methods in prompt compression on standard benchmark evaluations on both prompt-aware and prompt-agnostic compression setups and shows strong generalization across tasks and domains. To enable quick reproduction and further developments, we will release the datasets and the codebase upon publication of the paper.
\end{enumerate}

\section{Related Works}

In this section, we review the related works from two perspectives: prompt compression, and sentence embedding. 

\subsection{Prompt Compression}

Recent efforts in prompt compression focus on reducing the inference cost of LLMs. A key line of research involves model-agnostic methods, which leverage pre-trained models for compression. Early works like token pruning during the forward pass \citep{kim2022learned} and recursive context summarization \citep{chenwalking} introduced effective strategies but required access to the pre-trained LLM, which is often impractical. To address this limitation, newer approaches such as LLMLingua \citep{jiangllmlingua} utilize token-level perplexities to filter out semantically insignificant tokens, while Selective-Context \citep{li2023compressing} retains high-perplexity tokens using decoder-only models like LLaMa and GPT-2. LongLLMLingua \citep{jiang2023longllmlingua} further refines this idea by integrating question relevance for context-aware compression. However, these methods often lack adaptability to new domains, restricting their broader applicability.

In parallel, trainable methods for prompt compression have also gained traction as a key research direction. Soft prompt techniques \citep{wang2024adapting, bulatov2022recurrent, chevalier2023adapting} fine-tune or pre-train LLMs to achieve high compression rates, though they provide limited interpretability and control over the compression ratio. Sequence-to-sequence models compress context by generating a summary directly \citep{xu2024recomp}, but their autoregressive design introduces latency. Reinforcement learning approaches, such as optimizing for simplicity, fluency, and salience \citep{labankeep}, or using compression ratio as a reward \citep{jung2024discrete}, offer an alternative,  but may fall short in question-aware tasks. Recent innovations, such as \citep{pan2024llmlingua}, propose models that evaluate and prune tokens based on their information value, providing a more systematic approach to compression. Despite these advances, challenges remain in balancing efficiency, accuracy, and domain adaptability. More recently CPC \cite{CPC} proposed to utilize a context-aware sentence encoder to find the relevance of each sentence in the context to remove irrelevant sentences from the input prompt, achieving SOTA performance on existing benchmarks. However, a major limitation of such an approach is its reliance on the input question to guide the compression, requiring manual prompt formats for different tasks. While there have been some efforts towards building question-agnostic prompt compression \cite{pan2024llmlingua}, the performance of such methods falls short of the question-aware counterpart. Our goal in this work is to develop a question-agnostic prompt compression method without sacrificing performance.

\begin{figure*}
    \centering
    \includegraphics[width=0.65\linewidth]{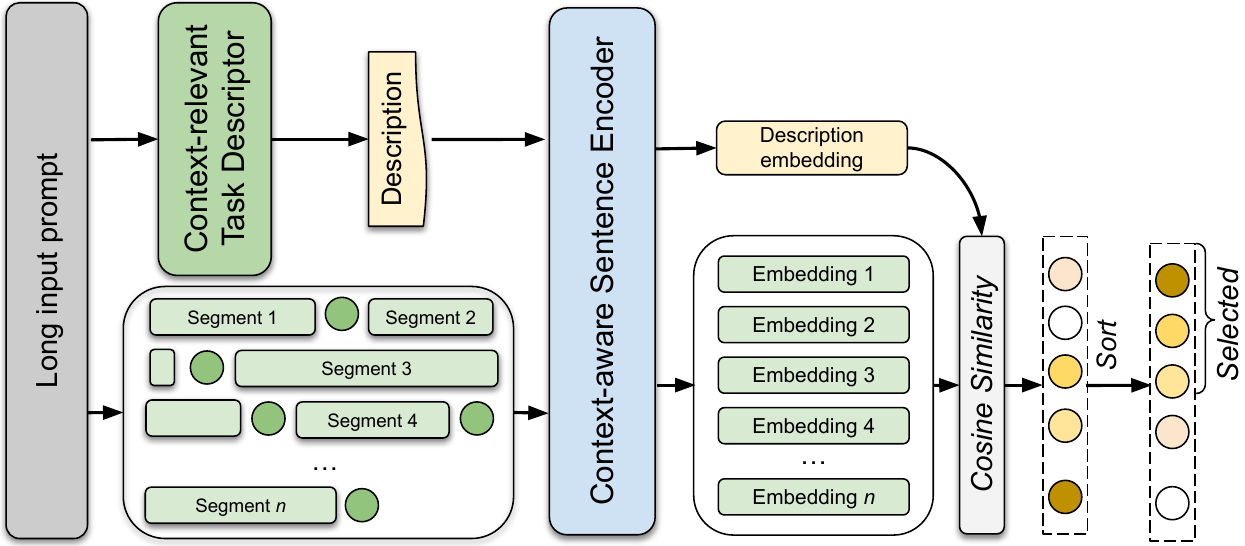}
    \caption{Illustration of our proposed prompt compression method. The CTD module generates a task description that is relevant to the input context. This description is then utilized by the Context-Aware Sentence Encoder to evaluate the relevance of each sentence in the input prompt, ultimately generating the compressed prompt.}
    \label{fig:main}
\end{figure*}

\subsection{Sentence Embedding Learning}
Text embedding learning aims to create high-dimensional vector representations of text that capture semantic meaning, facilitating tasks such as text classification, clustering, and similarity search. Early approaches like GloVe \citep{pennington2014glove} and Word2Vec \citep{church2017word2vec} focused on word- or token-level embeddings. Later, sentence-level representation learning gained attention, with works like \citet{reimers2019sentence} fine-tuning BERT-based models \citep{vaswani2017attention} to extract sentence embeddings for measuring text similarities. Subsequent research \citep{li2023towards, wang2022text, beltagy2020longformer} enhanced the effectiveness of these embeddings, particularly for longer contexts. Later, \citet{behnamghader2024llm2vec} leveraged the extensive knowledge of pre-trained LLMs to build robust sentence encoders. However, while these models excel at sentence representation, they lack context awareness, which is an essential property for our prompt compression approach. While a more recent work, CPC \citep{CPC} explored context-aware sentence encoding to compress prompts based on their relevance to the question, the proposed encoder was not task-agnostic. In this work, we introduce a task-agnostic and context-aware sentence encoder that captures the semantics of the context in its embedding.

\section{Proposed Method}

\subsection{Task-agnostic Prompt Compression}
The conventional setup of question-aware prompt compression operates by leveraging an explicit question (or a hand-crafted prompt where no question is available, e.g., code) $q$ provided along with the input context $c$. Specifically, this approach utilizes a sentence encoder $f_s$ to encode both the question and individual sentences $s_1,..,s_n$ for $n$ sentences from the context into embeddings $\mathbf{e}_q=f_s(q)$ and $\mathbf{e}_{s_1}..\mathbf{e}_{s_n}=f_s(s_1,...,s_n)$, respectively. The relevance of each sentence is then determined by a scoring function (e.g., cosine similarity) $\mathcal{R}(\mathbf{e}_q, \mathbf{e}_{s_i})$, allowing for the selection of the top $k$ most relevant sentences:
\begin{equation}\label{eq:compression}
    \mathcal{S} = \phi(q, c) = \operatorname{TopK}(\{\mathcal{R}(\mathbf{e}_q, \mathbf{e}_{s_i}) \mid s_i \in c\}),
\end{equation}
where $ \mathcal{S} $ is the compressed prompt containing the selected sentences. The encoder $f_s$ is trained on a dataset of $(q, c, pos, \{neg^{i=1..m}\})$ tuples to learn context-aware sentence embeddings, where $q$ is a question, $c$ is the context, and $pos$ is the positive sentence, and ${neg^{i=1..m}}$ are negative sentences. Here, a positive is a sentence containing relevant information to the question, while the negatives are context sentences containing no information regarding the question. While effective, this approach inherently relies on the availability of explicit questions and cannot be applied when no question is provided.

To address the limitations of question-aware compression, we propose Task-agnostic Prompt Compression (TPC), a novel solution for prompt compression that does not require such explicit questions. Instead, our approach generates a context-relevant task-description  $\hat{q}$ from the input prompt $p$ using our proposed Context-relevant Task Descriptor (CTD) model $ f_q $ as $\hat{q} = f_q(p; \theta_q)$. CTD is a lightweight causal LM designed to generate a task description that highlights the most relevant information within $p$. The task description $\hat{q}$ is then used to find the relevance of each sentence in the input prompt with our context-aware sentence encoder. 
To ensure the generated task description is semantically meaningful and contextually relevant, we train $f_q$ on a synthetic dataset of $(p, q)$ pairs (curated using our proposed data curation pipeline discussed in Section \ref{sec:CTD}), followed by a reward-guided fine-tuning stage with our proposed reward function (Section \ref{sec:reinforcement_learning_for_question_generation}). Finally, we propose a new method for training the Context-aware Sentence Encoder (CSE) that effectively captures the context in the embedding (Section \ref{sec:sen_enc}). The overall diagram of our proposed method is illustrated in Figure \ref{fig:main}, and a pseudo-code is provided in Algorithm \ref{alg:tpc} (Appendix A.1.).

\subsection{Context-relevant Task Descriptor}\label{sec:CTD}

CTD is an essential component of our proposed solution, which generates a context-aware task description $\hat{q}$ from the input prompt $p$. The CTD comprises two primary components: (1) a dataset curation pipeline for generating high-quality $(p, q)$ pairs, and (2) a causal encoder training process on the curated dataset to generate the task description $\hat{q}$. Below, we describe these components in detail.

\subsubsection{Dataset Curation}\label{sec:CTD_dataset}
To train the CTD, we require a dataset of $(p, q)$ pairs, where $p$ represents a long input prompt and $q$ is a relevant query or description that highlights the essential information in $p$. We leverage an existing pre-trained LLM to generate these pairs. Specifically, for a given input prompt $p$, the LLM is prompted to generate $q$ using the pre-designed Prompt 1 (see Appendix~\ref{app:method}). By applying this prompt to a dataset of long texts (e.g. Wikipedia), we generate a collection of question and input prompt pairs.

In real-world scenarios, user prompts are often structured, typically comprising an input prompt, a question, and an instruction. To ensure that our generated dataset adheres to such structure and remains suitable for downstream tasks, we refine the initially generated dataset by prompting the same pre-trained LLM with Prompt 2 (Appendix \ref{app:method}). This two-stage process yields a dataset $\mathcal{D}_{\text{CTD}}$ of input prompts and structured task descriptions, which is subsequently used to train the question generator $f_q$, as described below.

\subsubsection{Training CTD}\label{sec:question_generator}
At the core of CTD is a causal encoder, $f_q$, designed to generate contextually relevant task descriptions. We utilize a causal language model for $f_q$, initialized from a pre-trained LLM, to auto-regressively generate the task description $\hat{q}$ conditioned on the input prompt $p$. Formally, given input long prompt $p$, $f_q$ generates the question $\hat{q}$ as follows:
\vspace{-5pt}
\begin{equation}
    P(q \mid p; \theta_q) = \prod_{t=1}^{T} P(q_t \mid q_{<t}, p; \theta_q),
\end{equation}
where $q_t$ is the $t$-th token in $q$, $q_{<t}$ represents the sequence of tokens generated up to step $t-1$, and $\theta_q$ are the model parameters. We train the encoder $f_q$ to maximize the likelihood of generating the target question $q$ given the prompt $p$. The Supervised Training (ST) procedure is carried out using the following loss function:
\begin{equation}
\mathcal{L} = -\mathbb{E}_{(p, q) \sim \mathcal{D}_{\text{CTD}}'} \left[ \sum_{t=1}^{T} \log P(q_t \mid q_{<t}, p; \theta_q) \right],
\end{equation} where 
$T$ is the total number of tokens in $q$.

\subsection{Reward-guided Refinement}\label{sec:reinforcement_learning_for_question_generation}
As mentioned earlier, the generated task description serves as a guide for creating the compressed prompt by calculating the embedding similarity between the task description and each sentence in the prompt. Consequently, the quality of the compressed prompt is directly dependent on the generated task description. While the dataset $\mathcal{D}_{\text{CTD}}$ generated using the initial prompting pipeline provides a diverse set of $(p, q)$ pairs for training $f_q$, there is no explicit constraint to ensure that $\hat{q}$ generated by the LLM is the most relevant task description for input prompt $p$. 
To this end, we propose a novel reward function for fine-tuning $f_q$ through cross-entropy reinforcement learning. The reward function is specifically designed to achieve efficient prompt compression by generating a task description $\hat{q}$ that ensures the performance of the compressed prompt is on par with that of the original uncompressed input prompt.

\begin{figure}
    \centering
    \includegraphics[width=0.85\columnwidth]{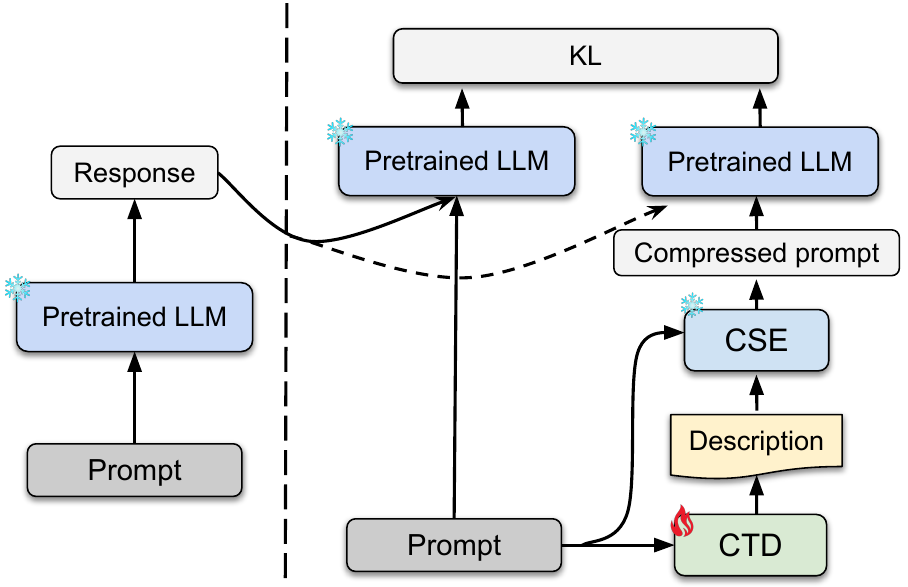}
    \caption{Overview of our proposed reward system for refining CTD with RL. (left) A response is generated by the pre-trained LLM with the complete long input prompt. (right) The CTD and CSE modules generate the compressed prompt. KL divergence between the conditional distribution of the generated response from the long prompt and the compressed prompt is used as the reward signal for the RL.}
    \label{fig:reward_system}
\end{figure}

Consider, task-descriptor $f_q$ as a model that performs the action of generating candidate task-description $\hat{q}$ from an input prompt $p$ of a real-world human instruction dataset (e.g. Tulu SFT dataset \cite{ivison2023camels}), which is then used to generate the compressed prompt $\mathcal{S}=\phi(\hat{q}, p)$ using Eq. \ref{eq:compression}. We define the reward function as the KL divergence between the conditional probability of generating the response $r$ from the whole input prompt $p$ and the compressed prompt $\mathcal{S}$ as:
\begin{equation}\label{eq:reward}
    R(q_i) = -\text{KL}(P(r \mid \mathcal{S}) || P(r \mid p))
\end{equation}
where, $r$ is the response from the pre-trained LLM using the whole input prompt $p$ as $r=f_{LLM}(p)$, and $P(r|\mathcal{S})$ is the conditional probability of generating the response $r$ given the compressed prompt $\mathcal{S}$ with a pre-trained LLM $f_{\text{LLM}}$,
\begin{equation}
    P(r \mid \mathcal{S}) = f_{\text{LLM}}(\mathcal{S}, r; \theta),
\end{equation}
and $P(r \mid p) $ is the conditional probability of generating the response $r $ given the original input prompt $p$. The KL divergence measures how much the distribution $P(r \mid \mathcal{S}) $ deviates from $P(r \mid p)$, which we use as the reward signal for encouraging the model to generate an informative task description $\hat{q}$ for compressing the prompt. Finally, 
using $f_q$ as an agent to generate multiple task descriptions $\hat{q}$ (actions), and calculating the corresponding reward, we fine-tune the task descriptor LM with the following loss:
\begin{equation}
    \mathcal{L}_{\text{RL}} = - \left[ \sum_{t=1}^{T} \log P(q_{j, t} \mid q_{j, {<t}}, p; \theta_q) \right],
\end{equation}
where $q_j$ is the generated task-description with maximal reward from Eq \ref{eq:reward}.  The overall process is illustrated in Figure \ref{fig:reward_system}.

\subsection{Context-aware Sentence Encoder}\label{sec:sen_enc}
CSE captures the information of the surrounding context in the embedding of a sentence, which is required for filtering out irrelevant sentences from the input prompt to generate the compressed prompt. Training CSE consists of two important components discussed below: data curation and encoder training. 

\subsubsection{Dataset Curation}
To train the context-aware sentence encoder, we require a dataset consisting of tuples containing long contexts, questions, positive sentence, and negative sentences. The context provides all the relevant information necessary to answer the question, positives are sentences within the context that contain partial but not necessarily complete information needed to answer the question, while negative sentences are context sentences that provide no relevant information. While a similar dataset (CQR) was introduced in CPC \cite{CPC}, the simple prompt used in the pipeline did not explicitly ensure that all the information is not present in the positive alone for encouraging understanding of the context. 
To this end, we introduce a multi-hop question-answer prompt for generating a new dataset of contexts, questions, positive sentence, and negative sentences using Prompt 3 (Appendix \ref{app:method}). The multi-hop question-answer prompt ensures that the generated questions require reasoning over multiple sentences in the context to arrive at an answer, thereby promoting deeper contextual understanding and ensuring that the positive does not contain the entire information required to answer the question. We denote the generated data of $(q, c, pos, \{neg^{i=1 \cdots m}\})$ tuples as the Multi-hop Context-Question Relevant dataset (MCQR), which is then used to train the CSE $f_s$.

\begin{table*}[!t]
    \centering
    \small
    \setlength{\tabcolsep}{1.5mm}
    \begin{tabular}{l|ccccccccc}
    \toprule
        \multirow{2}{*}{\textbf{Methods}} &  \multicolumn{8}{@{}c}{{\bf LongBench}} \\
        & SingleDoc & MultiDoc & Summ. & FewShot & Synth. & Code & Tokens & $1/\tau$\\
    \midrule
    \midrule
      \multicolumn{9}{@{}l}{{ \textit{Prompt-aware compression}}} \\ 
    Selective-Context & 29.74 & 43.86 & 23.61 & 57.95 & 22.75 & 45.32 & 1,925 & 5$\times$ \\
    LLMLingua & 20.48 & 37.02 & 21.23 & 55.99 & 10.5 & 43.78 & 1,856 & 6$\times$ \\
    LLMLingua-2 & 33.23 & 52.4 & 24.64 & 46.74 & 26.0 & 59.94 & 1,868 & 6$\times$ \\
    {LongLLMLingua} & 26.61 & 39.85 & 22.37 & 58.53 & 14.25 & 29.08 & 1,926 & 6$\times$ \\
    CompAct & 19.8 & 40.55 & - & 59.91 & - & - & 831 & 13$\times$ \\
    FAVICOMP & 27.13 & 51.65 & - & 41.63 & - & - & 95 & 120$\times$ \\
    CPC & 43.79 & 62.67 & {25.8} & 67.67 & \textbf{54.0} & 56.74 & 1,936 & 5$\times$ \\
    \cellcolor[rgb]{0.925,0.957,1}TPC-Base &\cellcolor[rgb]{0.925,0.957,1} 44.06 &\cellcolor[rgb]{0.925,0.957,1} 62.25 & \cellcolor[rgb]{0.925,0.957,1}25.66 & \cellcolor[rgb]{0.925,0.957,1}\textbf{70.1} & \cellcolor[rgb]{0.925,0.957,1}53.75 & \cellcolor[rgb]{0.925,0.957,1}59.72 & \cellcolor[rgb]{0.925,0.957,1}1,973 &\cellcolor[rgb]{0.925,0.957,1} 5$\times$ \\
    \cellcolor[rgb]{0.925,0.957,1}TPC-Large &\cellcolor[rgb]{0.925,0.957,1} 42.96 & \cellcolor[rgb]{0.925,0.957,1}58.38 & \cellcolor[rgb]{0.925,0.957,1}25.34 & \cellcolor[rgb]{0.925,0.957,1}69.34 & \cellcolor[rgb]{0.925,0.957,1}52.75 & \cellcolor[rgb]{0.925,0.957,1}58.9 & \cellcolor[rgb]{0.925,0.957,1}1,973 & \cellcolor[rgb]{0.925,0.957,1}5$\times$ \\
    \cellcolor[rgb]{0.925,0.957,1}TPC-Huge & \cellcolor[rgb]{0.925,0.957,1}\textbf{44.7} & \cellcolor[rgb]{0.925,0.957,1}\textbf{64.02} & \cellcolor[rgb]{0.925,0.957,1}\textbf{25.84} &\cellcolor[rgb]{0.925,0.957,1} 68.91 &\cellcolor[rgb]{0.925,0.957,1} \textbf{54.0} &\cellcolor[rgb]{0.925,0.957,1} \textbf{60.04} &\cellcolor[rgb]{0.925,0.957,1} 1,970 & \cellcolor[rgb]{0.925,0.957,1}5$\times$ \\
    \cmidrule (lr){1-9}
    \multicolumn{7}{@{}l}{{ \textit{Prompt-agnostic compression}}} \\
    Selective-Context & 14.72 & 13.08 & 22.04 & 21.09 & 1.0 & 19.3 & 1,756 & 6$\times$ \\
    LLMLingua & 21.02 & 34.79 & 21.32 & {54.57} & 7.0 & \textbf{64.62} & 1,870 & 6$\times$ \\
    LLMLingua-2 & 20.18 & 7.0 & 22.86 & 13.02 & 1.34 & 23.3 & 1,866 & 6$\times$ \\
    \cellcolor[rgb]{0.925,0.957,1}TPC-Base & \cellcolor[rgb]{0.925,0.957,1}40.7 & \cellcolor[rgb]{0.925,0.957,1}58.23 & \cellcolor[rgb]{0.925,0.957,1}25.66 & \cellcolor[rgb]{0.925,0.957,1}55.95 &\cellcolor[rgb]{0.925,0.957,1} 27.38 & \cellcolor[rgb]{0.925,0.957,1}\textbf{38.36} & \cellcolor[rgb]{0.925,0.957,1}1,934 & \cellcolor[rgb]{0.925,0.957,1}5$\times$ \\ 
    \cellcolor[rgb]{0.925,0.957,1}TPC-Large & \cellcolor[rgb]{0.925,0.957,1}41.59 & \cellcolor[rgb]{0.925,0.957,1}57.59 & \cellcolor[rgb]{0.925,0.957,1}\textbf{26.26} & \cellcolor[rgb]{0.925,0.957,1}47.86 & \cellcolor[rgb]{0.925,0.957,1}\textbf{42.0} & \cellcolor[rgb]{0.925,0.957,1}35.25 & \cellcolor[rgb]{0.925,0.957,1}1,978 & \cellcolor[rgb]{0.925,0.957,1}5$\times$ \\
    \cellcolor[rgb]{0.925,0.957,1}TPC-Huge & \cellcolor[rgb]{0.925,0.957,1}\textbf{41.84} & \cellcolor[rgb]{0.925,0.957,1}\textbf{61.93} & \cellcolor[rgb]{0.925,0.957,1}26.22 & \cellcolor[rgb]{0.925,0.957,1}\textbf{58.27} & \cellcolor[rgb]{0.925,0.957,1}39.0 & \cellcolor[rgb]{0.925,0.957,1}36.25 & \cellcolor[rgb]{0.925,0.957,1}1,962 & \cellcolor[rgb]{0.925,0.957,1}5$\times$ \\

    \bottomrule
    \end{tabular}
    \caption{Performance of different methods in prompt-aware and prompt-agnostic setups on LongBench. 
    }
    \label{tab:main_result_long_context}
\end{table*}

\subsubsection{Encoder Training}

CSE is trained to learn the context-aware sentence representations by distinguishing the positives and negatives. Specifically, this loss maximizes the similarity between the embeddings of the question and the positive sentence while minimizing the similarity between the question and the embeddings of negative sentences. Similar to CPC \cite{CPC}, we use a contrastive loss along with the MNTP loss \cite{behnamghader2024llm2vec} for more stable training on our newly curated MCQR dataset, with another key modification. Specifically, we introduce two new types of tokens into the tokenizer dictionary: the end-of-sentence token \texttt{<end\_of\_sent>} and the question token \texttt{<end\_of\_question>}. The \texttt{<end\_of\_sent>} token is inserted at the end of each particular sentence in the text, explicitly marking sentence boundaries and enabling the model to process text segments of varying lengths effectively. This token enables the adaptation to different input granularities while maintaining a clear structural understanding of the text. Similarly, the \texttt{<end\_of\_question>} token is appended to the end of each question, signalling its conclusion and helping the model handle questions with diverse syntactic structures. Together, these tokens improve the model's ability to generate embeddings that encode contextual and structural relationships more effectively.

\section{Experiments}
In this section, we discuss the experiments and results of our proposed solution for prompt compression. We begin by presenting the datasets, evaluation protocols, and implementation details. Subsequently, we discuss the main results, ablation studies, and qualitative findings on TPC.

\subsection{Datasets}\label{sec:dataset}

\textbf{Evaluation datasets.} To evaluate the performance of TPC, we adhere to the experimental protocols outlined in prior research \cite{jiang2023longllmlingua} and benchmark our method on \textbf{LongBench} \cite{bai2023longbench} and \textbf{ZeroSCROLLS} \cite{shaham2023zeroscrolls} datasets. LongBench features a diverse range of tasks, including both single-document and multi-document QA, summarization, few-shot learning, synthetic and code generation. Similarly, ZeroSCROLLS offers a collection of tasks including GovReport, SummScreenFD,
QMSUM, SQuality, Quality, NarrativeQA, Qasper, Musique, SpaceDigest, BookSumSort, and Tokens.

\textbf{Our curated datasets.} As outlined in the Methods section, TPC is trained using two of our curated datasets: CTD and MCQR. The CTD dataset comprises 4,100 context-question pairs specifically designed for training the CTD module. The MCQR dataset includes 9,300 samples, each consisting of a question, a context, a positive and negative sentences.

\subsection{Evaluation Protocols}
We conduct a series of experiments across multiple downstream tasks, following widely accepted protocols on the LongBench and ZeroSCROLLS subsets. Our approach adheres to the standard evaluation protocols established in prior studies \cite{jiang2023longllmlingua}. Specifically, for summarization tasks, we assess our method by comparing the Rouge scores \citep{lin2004rouge} between the ground truth responses and the outputs generated by the model using compressed prompts. In document QA tasks, the model's responses are evaluated against the ground truth using the F1 score. For code completion tasks, we rely on a textual similarity measure derived from the Levenshtein edit distance \citep{yujian2007normalized}. 

We present the main results in two evaluation setups, including prompt-aware compression and prompt-agnostic compression. Prompt-aware compression is the most common setup adopted by existing works where the long context is provided with a question, and the compression is performed based on the question. However, this limits its applicability to certain tasks such as summarization and code, where a manual hand-designed and task-specific prompt template is required for the compression. On the contrary, prompt-agnostic compression does not require such a template or an explicit question to compress the context.

\begin{table*}[!t]
    \centering
    \small
    \setlength{\tabcolsep}{1.2mm}
    \begin{tabular}{l|cccccccccccc}
    \toprule
        \multirow{8}{*}{\textbf{Methods}} &  \multicolumn{11}{@{}c}{{\bf ZeroSCROLLs}} \\
        & \rotatebox{70}{GovReport} & \rotatebox{70}{SummScreenFD} & \rotatebox{70}{QMSUM} & \rotatebox{70}{SQuality} & \rotatebox{70}{Quality} & \rotatebox{70}{NarrativeQA} & \rotatebox{70}{Qasper} & \rotatebox{70}{Musique} & \rotatebox{70}{SpaceDigest} & \rotatebox{70}{BookSumSort} & \rotatebox{70}{Tokens} & \rotatebox{70}{$1/\tau$} \\
    \midrule
    \midrule
      \multicolumn{10}{@{}l}{{ \textit{Prompt-aware compression}}} \\ 
    Selective-Context & 17.92 & 9.91 & 12.43 & 15.57 & 76.19 & 33.48 & 32.78 & 29.76 & 39.52 & 75.33 & 1,925 & 5$\times$ \\
    LLMLingua & 15.03 & 6.52 & 11.03 & 12.14 & 71.43 & 23.08 & 15.85 & \textbf{35.00} & 43.15 & 56.17 & 1,856 & 6$\times$ \\
    LLMLingua-2 & 17.60 & \textbf{12.73} & 15.25 & \textbf{17.37} & 80.95 & 34.75 & 33.87 & 30.00 & 37.20 & 64.98 & 1,868 & 6$\times$ \\
    {LongLLMLingua} & 0.0 & 0.0 & 0.0 & 14.45 & 71.43 & 21.6 & 22.81 & 0.0 & 0.0 & 0.0 & 1,926 & 6$\times$ \\
    CompAct & - & - & 11.64 & 13.1 & 66.67 & 17.75 & 4.42 & 20.0 & - & - & 831 & 13$\times$ \\
    FAVICOMP & - & - & 13.14 & 13.0 & 42.86 & 35.86 & 24.05 & 31.76 & - & - & 95 & 120$\times$ \\
    CPC & 18.67 & 12.44 & \textbf{17.06} & 16.49 & \textbf{100.0} & \textbf{45.84} & 41.24 & 29.76 & 54.36 & 80.51 & 1,936 & 5$\times$ \\
    \cellcolor[rgb]{0.925,0.957,1}TPC-Base & \cellcolor[rgb]{0.925,0.957,1}18.67 & \cellcolor[rgb]{0.925,0.957,1}12.26 & \cellcolor[rgb]{0.925,0.957,1}15.78 & \cellcolor[rgb]{0.925,0.957,1}17.00 & \cellcolor[rgb]{0.925,0.957,1}90.48 & \cellcolor[rgb]{0.925,0.957,1}44.75 & \cellcolor[rgb]{0.925,0.957,1}\textbf{41.58} & \cellcolor[rgb]{0.925,0.957,1}29.76 & \cellcolor[rgb]{0.925,0.957,1}53.90 & \cellcolor[rgb]{0.925,0.957,1}80.36 & \cellcolor[rgb]{0.925,0.957,1}1,973 &\cellcolor[rgb]{0.925,0.957,1} 5$\times$ \\
    \cellcolor[rgb]{0.925,0.957,1}TPC-Large & \cellcolor[rgb]{0.925,0.957,1}19.30 & \cellcolor[rgb]{0.925,0.957,1}12.14 & \cellcolor[rgb]{0.925,0.957,1}16.38 & \cellcolor[rgb]{0.925,0.957,1}16.64 & \cellcolor[rgb]{0.925,0.957,1}\textbf{100.00} & \cellcolor[rgb]{0.925,0.957,1}40.75 & \cellcolor[rgb]{0.925,0.957,1}40.65 & \cellcolor[rgb]{0.925,0.957,1}29.76 & \cellcolor[rgb]{0.925,0.957,1}61.42 & \cellcolor[rgb]{0.925,0.957,1}83.33 & \cellcolor[rgb]{0.925,0.957,1}1,973 &\cellcolor[rgb]{0.925,0.957,1} 5$\times$ \\
    \cellcolor[rgb]{0.925,0.957,1}TPC-Huge & \cellcolor[rgb]{0.925,0.957,1}\textbf{19.30} & \cellcolor[rgb]{0.925,0.957,1}12.14 & \cellcolor[rgb]{0.925,0.957,1}16.38 & \cellcolor[rgb]{0.925,0.957,1}16.64 & \cellcolor[rgb]{0.925,0.957,1}\textbf{100.00} & \cellcolor[rgb]{0.925,0.957,1}40.75 & \cellcolor[rgb]{0.925,0.957,1}41.35 & \cellcolor[rgb]{0.925,0.957,1}29.76 & \cellcolor[rgb]{0.925,0.957,1}\textbf{61.42} & \cellcolor[rgb]{0.925,0.957,1}\textbf{83.33} &\cellcolor[rgb]{0.925,0.957,1} 1,970 &\cellcolor[rgb]{0.925,0.957,1} 5$\times$ \\
    \cmidrule (lr){1-13}
    \multicolumn{7}{@{}l}{{ \textit{Prompt-agnostic compression}}} \\
    Selective-Context & 17.94 & 10.71 & 7.96 & 13.14 & 33.33 & 1.14 & 25.58 & 33.33 & 19.31 & \textbf{76.5} & 1,756 & 6$\times$ \\
    LLMLingua & 13.75 & 7.37 & 10.15 & 10.56 & 28.57 & 12.56 & 21.50 & 20.0 & 34.53 & 24.54 & 1,870 & 6$\times$ \\
    LLMLingua-2 & 18.24 & \textbf{13.09} & 11.07 & 16.33 & 38.10 & 1.11 & \textbf{26.74} & \textbf{39.76} & 29.49 & 63.63 & 1,866 & 6$\times$ \\
    \cellcolor[rgb]{0.925,0.957,1}TPC-Base & \cellcolor[rgb]{0.925,0.957,1}\textbf{19.72} &\cellcolor[rgb]{0.925,0.957,1} \textbf{13.07} &\cellcolor[rgb]{0.925,0.957,1} \textbf{16.94} & \cellcolor[rgb]{0.925,0.957,1}\textbf{17.29} &\cellcolor[rgb]{0.925,0.957,1} \textbf{90.91} &\cellcolor[rgb]{0.925,0.957,1} {19.97} &\cellcolor[rgb]{0.925,0.957,1} 16.54 &\cellcolor[rgb]{0.925,0.957,1} 33.11 & \cellcolor[rgb]{0.925,0.957,1}58.73 & \cellcolor[rgb]{0.925,0.957,1}23.00 &\cellcolor[rgb]{0.925,0.957,1} 1,978 &\cellcolor[rgb]{0.925,0.957,1} 5$\times$ \\
    \cellcolor[rgb]{0.925,0.957,1}TPC-Large & \cellcolor[rgb]{0.925,0.957,1}18.48 &\cellcolor[rgb]{0.925,0.957,1} 12.3 &\cellcolor[rgb]{0.925,0.957,1} 16.02 & \cellcolor[rgb]{0.925,0.957,1}\textbf{17.29} &\cellcolor[rgb]{0.925,0.957,1} \textbf{90.91} &\cellcolor[rgb]{0.925,0.957,1} \textbf{21.6} &\cellcolor[rgb]{0.925,0.957,1} \textbf{26.7} &\cellcolor[rgb]{0.925,0.957,1} 33.29 & \cellcolor[rgb]{0.925,0.957,1}59.51 & \cellcolor[rgb]{0.925,0.957,1}5.24 &\cellcolor[rgb]{0.925,0.957,1} 1,978 &\cellcolor[rgb]{0.925,0.957,1} 5$\times$ \\
    \cellcolor[rgb]{0.925,0.957,1}TPC-Huge & \cellcolor[rgb]{0.925,0.957,1}17.73 &\cellcolor[rgb]{0.925,0.957,1} 12.4 &\cellcolor[rgb]{0.925,0.957,1} 16.28 & \cellcolor[rgb]{0.925,0.957,1}{17.26} &\cellcolor[rgb]{0.925,0.957,1} 86.36 &\cellcolor[rgb]{0.925,0.957,1} 16.42 &\cellcolor[rgb]{0.925,0.957,1} 15.19 &\cellcolor[rgb]{0.925,0.957,1} 24.62 & \cellcolor[rgb]{0.925,0.957,1}\textbf{63.2} & \cellcolor[rgb]{0.925,0.957,1}48.56 &\cellcolor[rgb]{0.925,0.957,1} 1,978 &\cellcolor[rgb]{0.925,0.957,1} 5$\times$ \\

    \bottomrule
    \end{tabular}
    \caption{Performance of different methods in prompt-aware and prompt-agnostic setups on ZeroSCROLLs. 
    }
    \label{tab:zero_scrolls_result_long_context}
    \vspace{10pt}
\end{table*}

\begin{table*}[t]
\hspace{-35pt}
\small
    \centering
    \subfloat[
    Ablation on CTD.
    \label{tab:ablation_ds}
    ]{
    \hspace{1pt}
    \begin{minipage}{0.24\linewidth}{\begin{center}
    \tablestyle{2pt}{1.0}
        \begin{tabular}{l c}
        \toprule
        \textbf{Ablation} & \textbf{Perf.} \\
        \midrule
        TPC & 41.62\\
        TPC w/o RL &  37.97 \\
        TPC w/o ST and RL & 25.92\\
        \bottomrule
        \end{tabular}
    \end{center}}\end{minipage}
    }
    \centering
    \hspace{4em}
    \subfloat[
     Ablation on CSE.
    \label{tab:ablation_loss}
    ]{
    \centering
    \begin{minipage}{0.24\linewidth}{\begin{center}
    \tablestyle{2pt}{1.0}
        \begin{tabular}{l c}
        \toprule
        \textbf{Ablation} & \textbf{Perf.} \\
        \midrule
        CSE  & 51.28 \\
        CSE w/o MCQR & 49.7 \\
        CSE w/o spec. tokens & 50.23 \\ 
        
        \bottomrule
        \end{tabular}
    \end{center}}\end{minipage}
    }
    \hspace{3em}
    \subfloat[
    Performance vs. context length.
    \label{tab:ablation_sentence_embed}
    ]{
    \begin{minipage}{0.24\linewidth}{\begin{center}
    \tablestyle{2pt}{1.0}
    \begin{tabular}{l c c c c}
    \toprule
        \textbf{Setup} & \textbf{TPC-B}  & \textbf{TPC-L} & \textbf{TPC-H} & \textbf{CPC}  \\
        \midrule
        2k & 52.59 & 51.28 & 52.92 & 51.78\\
        3k  & 53.14 & 52.78 & 53.69 & 52.71\\ 
        \\
    \bottomrule
    \end{tabular}
    \end{center}}\end{minipage}
    }
\caption{Ablation and analysis of different components of TPC on LongBench. In (c), B, L, and H represent Base, Large, and Huge, respectively.
}
\label{tab:ablations}
\end{table*}

\begin{figure*}
    \centering
    \includegraphics[width=0.8\linewidth]{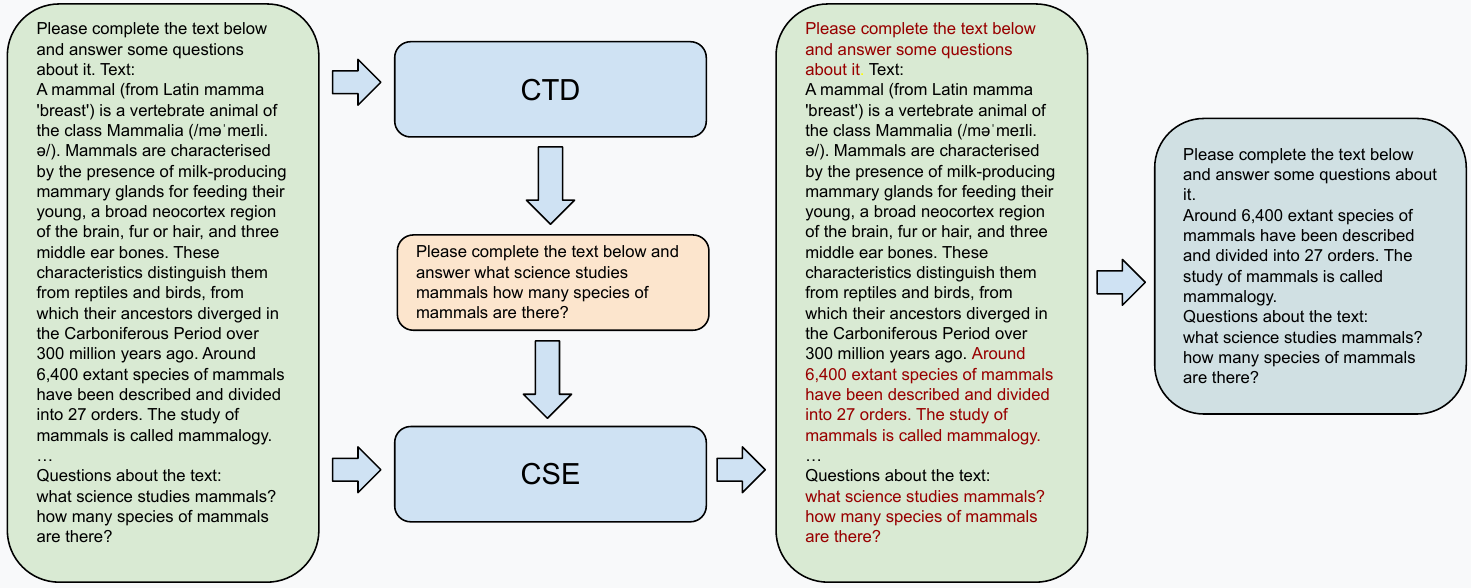}
    \caption{A qualitative example of prompt compression with TPC.}
    \label{fig:example}
    \vspace{-5pt}
\end{figure*}

\subsection{Implementation Details}
We train the $f_q$ with our curated CTD dataset for 2 epochs with an AdamW optimizer, a learning rate of 1.5e-4, and a batch size of 16. This is followed by the reward-guided RL training for 3 iterations with Llama-3.1-8B \cite{dubey2024llama} as the pre-trained encoder for the reward function. CSE is trained with an AdamW optimizer for 2 epochs, a learning rate of 5e-5, and a batch size of 32.
We introduce three versions of TPC, namely Base, Large, and Huge, consisting of Qwen2 \cite{yang2024qwen2} with 0.5B parameters, Llama-3.2-Instruct \cite{dubey2024llama} with 1B parameters, and Mistral-Instruct-v0.2 \cite{jiang2023mistral} with 7B parameters as the encoder, respectively. We initialize the encoder with the pre-trained weights and fine-tune it with LoRA \cite{hu2021lora} of rank 16, and follow the training details from \cite{behnamghader2024llm2vec} to turn the pre-trained causal encoder into bidirectional sentence encoder. All the experiments are conducted on an Nvidia A100 80GB GPU.

\subsection{Main Results}
We present the main results of our experiments on LongBench in Table \ref{tab:main_result_long_context} on both prompt-aware and prompt-agnostic compression setups. We perform the evaluations on all three sizes of our model (TPC-Base, TPC-Large, and TPC-Huge). As we find from this table, in the prompt-aware setup, TPC outperforms all the prior works across the tasks, with up to 3.3 points improvement on individual tasks. While being considerably smaller than the existing SOTA (CPC), our smallest model (TPC-Base) outperforms it on most setups. 

Similarly, TPC shows strong performance on the prompt-agnostic compression setup. Here, only a few of the existing methods can be adopted to the prompt-agnostic setup, due to the nature of the compression methods. In the prompt-agnostic setup, TPC outperforms existing methods by a larger margin than the prompt-aware setup. Specifically, TPC shows up to 35.0 points improvement (on synthetic task) over the existing methods, along with 20.82 points improvement on SingleDoc, and 27.14 points improvement on MultiDoc tasks. Despite having 14$\times$ less parameters than our largest model, our Base model achieves strong results, and considerably outperforms existing methods. 

The results on the ZeroSCROLLs dataset are presented in Table \ref{tab:zero_scrolls_result_long_context}. Similar to LongBench, the results are presented on both prompt-aware and prompt-agnostic compression setups for the three different model sizes. As we find from this table, on the prompt-aware setup, TPC outperforms prior works by up to 7.06 points (SpaceDigest). Notably, TPC outperforms existing methods by a larger margin on the prompt-agnostic setup. Specifically, we observe up to 51.81 points improvement on the Quality task. Additionally, we observe 28.67 and 9.04 points improvements for SpaceDigest and NarrativeQA tasks, respectively. Notably, our base encoder performs comparable to that of the large encoder, surpassing it in a few tasks, possibly due to the relative simplicity of the tasks in ZeroSCROLLs dataset.

\subsection{Ablation}
Here, we present ablation and sensitivity experiments on our proposed TPC. In Table \ref{tab:ablations}(a) we study the impact of different components of the CTD module, including the supervised fine-tuning and the reward-guided refinement. As we find from this experiment, removing the reward-guided refinement module shows a 3.65 point drop in the performance, while removing the ST module shows another 12.05 points drop in the overall performance, showing the importance of both modules in the final performance of the model. Note that RL is dependent on the ST module, and therefore can not be ablated alone.

Next, we ablate our two novel components in CSE, presented in Table \ref{tab:ablations}(b). First, we remove the new multi-hop dataset MCQR and train CSE with CQR instead. This results in a 1.58-point drop in performance, indicating the impact of our curated MCQR dataset over the existing CQR dataset for training a context-aware sentence encoder. Then, we remove the special tokens (\texttt{<end\_of\_sent>} and \texttt{<end\_of\_question>}) from the encoder, leading to a 1.05-point drop in performance. These experiments demonstrate the importance of each component in our context-aware sentence encoder. 

Next, in Table \ref{tab:ablations}(c), we present additional studies on different constraints for context length, including 2,000 and 3,000 tokens, and compare them with the prior SOTA CPC. From this study, we observe that TPC outperforms CPC under both constraint setups, where TPC-Huge outperforms CPC with a larger margin on the 2K constraint compared to the 3K token constraint.

\begin{figure}
    \centering
    \includegraphics[width=0.95\columnwidth]{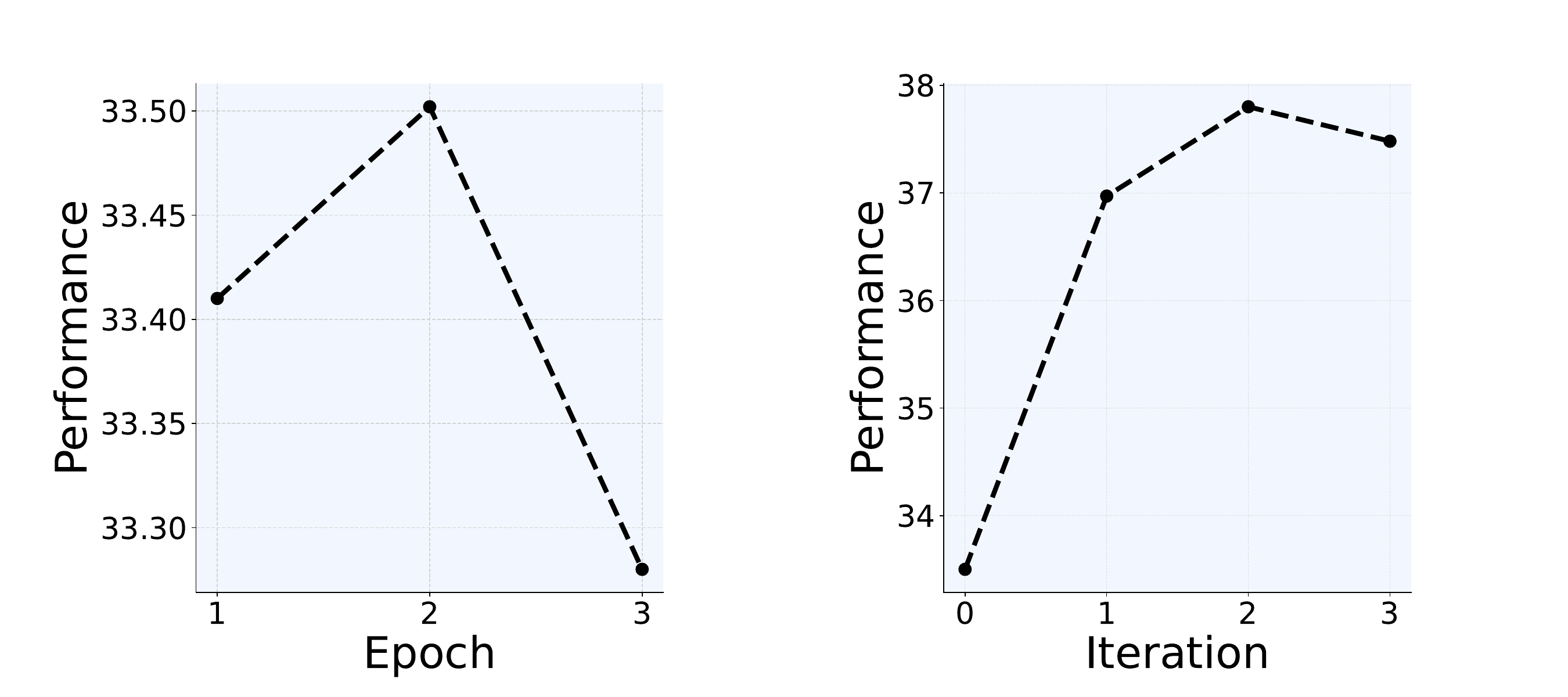}
    \caption{Sensitivity study on CTD training epochs (left) and the number of RL iterations  (right). Here, an RL iteration of 0 indicates no reward-guided refinement training.}
    \label{fig:rl_epochs}
    \vspace{-15pt}
\end{figure}

\subsection{Qualitative Analysis}
A qualitative example of the prompt compression pipeline of TPC is illustrated in Figure \ref{fig:example}. Here, the CTD module first generates a relevant task description from the input prompt, which is processed with the CSE block along with the input prompt to generate the compressed prompt.

\subsection{Sensitivity Analysis}
Finally, we investigate two critical hyper-parameters of our method: the number of epochs for training the CTD model and the number of iterations for fine-tuning the CTD using RL. The number of training epochs is crucial to ensure that the task descriptions generated by the CTD model are sufficiently accurate for the subsequent reward-guided fine-tuning stage. As shown in Figure \ref{fig:rl_epochs} (left), the model's performance drops after 2 epochs of training due to overfitting. 
The impact of the number of RL iterations is illustrated in Figure \ref{fig:rl_epochs} (right). Here, iteration 0 represents the baseline performance without RL fine-tuning. As we find from this figure, the performance initially improves with the number of iterations. However, the performance drops after 2 iterations likely due to overfitting to the recent policies.

\section{Conclusion}
We introduce TPC, a general-purpose prompt compression method that, unlike existing methods doesn't require an explicit question or handcrafted prompt for generating the compressed prompt. Instead, our method uses a task descriptor to generate a context-relevant task description, which is then utilized to find the relevance of each sentence in the prompt to produce the compressed prompt using CSE. 
Experiments on LongBench and ZeroSCROLLS demonstrate that our approach outperforms existing methods in both prompt-aware and prompt-agnostic compression settings. While achieving strong performance, our method is computationally efficient compared to existing SOTA. Our smallest variant with considerably fewer parameters performs on par with the previous SOTA.

\clearpage

\bibliography{icml25}
\bibliographystyle{icml2025}

\clearpage

\newpage
\appendix
\section{Appendix.}

\subsection{Method} \label{app:method}
In this section, we provide additional details on our proposed TPC method. First, we describe the prompts used in the training pipeline. Specifically, Prompt 1 is employed in the data curation pipeline for the CTD model, while Prompt 2 is utilized for curating the generated data. Finally, Prompt 3 is applied for the data curation process of the context-aware sentence encoder.

\noindent\fbox{\begin{minipage}{23em}
{\small
\noindent \texttt{\textbf{Prompt 1:}\\
You are tasked with writing user queries based on a long context. Consider the topic of the context when formulating the query. Queries should be concise and specific. You may also ask more complex questions, such as requesting specific knowledge extraction from the text, extending the text, answering questions based on the text, paraphrasing/rewriting the text, or summarizing the text. \\
Do not include the word "context" in the query. \\
Long context: \{text\} \\
Query: \\
}}
\end{minipage}}
\\

\noindent\fbox{\begin{minipage}{23em}
{\small
\noindent \texttt{\textbf{Prompt 2:}\\
You are tasked with writing a user query based on a long context. Create a complex template that integrates this context and a question into a single instruction.\\
The template must include the placeholders \texttt{\{text\}} and \texttt{\{question\}}.\\
Template examples: \\
1. You are given a text and a question related to it. Answer the question based on the text.\\
Question: \{question\}\\
Text: \{text\}\\
Now answer the question:\\
2. Can you extend the following block of code: \{text\}\\
such that it satisfies the requirement: \{question\}\\
Now write the code:\\
Provide various templates, taking into account the topic of the text and the question.
}}
\end{minipage}}
\\

\noindent\fbox{\begin{minipage}{23em}
{\small
\noindent \texttt{\textbf{Prompt 3:}\\
You are given a long text consisting of numbered sentences. Your task is to generate complex multi-hop questions about this text, such that answering them requires step-by-step reasoning (in multiple hops). To achieve this, first, ask a series of sequential factual questions and identify the corresponding sentences that contain the answers to these questions. Then, formulate a final question that can only be answered by combining the information from the previously generated questions and answers. Additionally, specify which sentences (by their numbers) contain the information necessary to answer the final question.\\
\\
Toy example:
[[1]] John is married to Mary. [[2]] They've decided to spend their marriage anniversary in Spain. [[3]] Mary was afraid that their two small children, Jody and Sue, were too small for a flight. [[4]] That's why she asked her elder sister Jane to look after them.\\
Questions:\\
Question 1: Who is John married to?\\
Answer 1: John is married to Mary, as stated in [[1]].\\
Question 2: How many children does Mary have?\\
Answer 2: Mary has two children, as stated in [[3]].\\
\\
Combining the questions to create a multi-hop question:\\
1. John is married to Mary, as stated in [[1]].\\
2. Mary has two children, as stated in [[3]].\\
Final question: How many children does John have?\\
Necessary sentences: [[1]], [[3]]\\
\\
Now, solve this example:\\
\{text\}\\
Questions:
}}
\end{minipage}}
\\

\clearpage

Next, we provide the pseudo-code of the overall pipeline of our proposed prompt compression method TPC in Algorithm \ref{alg:tpc}, the pseudo-code for context-relevant task descriptor in Algorithm \ref{alg:CTD}, and CTD refinement with RL in Algorithm \ref{alg:rl_CTD}.

\begin{algorithm}[t]
   \caption{Task-Agnostic Prompt Compression (TPC)}
   \label{alg:tpc}
\begin{algorithmic}
   \STATE {\bfseries Input:} Prompt $p$, context $c$, number of top sentences $k$, context-aware sentence encoder $f_s$, context-relevant task descriptor LM, $f_q$. 
   \STATE {\bfseries Output:} Compressed prompt $\mathcal{S}$
   \STATE \textcolor{gray}{// Generate task description $\hat{q}$ from prompt $p$ using $f_q$}
   \STATE $\hat{q} = f_q(p; \theta_q)$
   \STATE \textcolor{gray}{// Encode task description $\hat{q}$ and each sentence $s_i$ in context $c$ using sentence encoder $f_s$}
   \STATE  $\mathbf{e}_{\hat{q}} = f_s(\hat{q})$, \{$\mathbf{e}_{s_1},...,\mathbf{e}_{s_n}\} = f_s(s_1,..,s_n), \forall s_i \in c$
   \STATE  \textcolor{gray}{// Compute relevance scores for all sentences in $c$}
   \STATE $\mathcal{R}_i = \mathcal{R}(\mathbf{e}_{\hat{q}}, \mathbf{e}_{s_i}), \forall s_i \in c$
   \STATE  \textcolor{gray}{// Select top-$t$ sentences based on relevance scores}
   \STATE $\mathcal{S} = \operatorname{TopK}(\{\mathcal{R}_i \mid s_i \in c\}, k)$
   \STATE Return compressed prompt $\mathcal{S}$
\end{algorithmic}
\end{algorithm}

\begin{algorithm}[t]
   \caption{Context-relevant Task Descriptor (CTD)}
   \label{alg:CTD}
\begin{algorithmic}
   \STATE { \bfseries Input:} Prompt $p$, pre-trained LLM $g$, dataset $\mathcal{D}$, training parameters $\theta_q$.
   \STATE {\bfseries Output:} Context-relevant Question Generator $f_q$
   \STATE \textcolor{gray}{// Stage 1: Dataset Curation}
   \STATE   Initialize empty dataset $\mathcal{D}_{\text{raw}}$
   \FOR{each prompt $p \in \mathcal{D}$}
   \STATE Generate raw question $q$ using $g$: $q = g(p; \text{Prompt 1})$
   \STATE Append $(p, q)$ to $\mathcal{D}_{\text{raw}}$
   \ENDFOR
   \STATE \textcolor{gray}{// Stage 2: Structured Prompt Creation}
   \STATE Initialize empty dataset $\mathcal{D}_{\text{CTD}}$
   \FOR{each $(p, q) \in \mathcal{D}_{\text{raw}}$}
   \STATE  Generate structured prompt $p_{\text{struct}}$ using $g$: $p_{\text{struct}} = g(p, q; \text{Prompt 2})$
   \STATE  Append $(p_{\text{struct}}, q)$ to $\mathcal{D}_{\text{CTD}}$
   \ENDFOR
   \STATE   \textcolor{gray}{// Train Causal Encoder $f_q$}
   \STATE   Initialize $f_q$ with pre-trained weights $\theta_q$
   \REPEAT
   \STATE   Sample batch $(p_{\text{struct}}, q)$ from $\mathcal{D}_{\text{CTD}}$
   \STATE   Compute loss: 
   \STATE \quad $\mathcal{L}_{\text{CTD}} = -\sum_{t=1}^T \log P(q_t \mid q_{<t}, p_{\text{struct}}; \theta_q)$
   \STATE Update parameters $\theta_q$ using gradient descent
   \UNTIL{convergence criteria met}
   \STATE  Return trained question generator $f_q$
\end{algorithmic}
\end{algorithm}

\begin{algorithm}[t]
   \caption{Reinforcement Learning for CTD Refinement}
   \label{alg:rl_CTD}
\begin{algorithmic}
   \STATE { \bfseries Input:} Pre-trained task descriptor $f_q$, candidate task descriptions $\hat{\{q_i\}}$, pre-trained LLM $f_{\text{LLM}}$, input prompt $p$, response $r$
   \STATE {\bfseries Output:} Refined task descriptor $f_q$ with updated parameters $\theta_q$
   \STATE \textcolor{gray}{// Generate candidate task descriptions using $f_q$}
   \STATE $\hat{\{q_i\}} = f_q(p; \theta_q)$
   \STATE \textcolor{gray}{// For each $q_i$, construct compressed prompt $\mathcal{S}_i$ using $\phi$}
   \STATE $\mathcal{S}_i = \phi(q_i, p) \quad \forall q_i \in \hat{\{q_i\}}$
   \STATE \textcolor{gray}{// Compute initial response $r$ using $f_{\text{LLM}}$ and the initial non-compressed prompt $p$}
   \STATE $r = f_{\text{LLM}}(p)$
   \STATE \textcolor{gray}{// Evaluate candidate task descriptions based on KL divergence reward}
   \STATE $R(q_i) = -\text{KL}(P(r \mid \mathcal{S}_i) || P(r \mid p)) \quad \forall q_i \in \hat{\{q_i\}}$
   \STATE \textcolor{gray}{// Optimize $f_q$ using cross entropy RL for the generated task-description with maximal reward $q_j$}
   \STATE $\mathcal{L}_{\text{RL}} = - \left[ \sum_{t=1}^{T} \log P(q_{j, t} \mid q_{j, {<t}}, p; \theta_q) \right]$
   \STATE Update $\theta_q$ to minimize $\mathcal{L}_{\text{RL}}$
   \STATE Return refined $f_q$
\end{algorithmic}
\end{algorithm}

\subsection{Sample result}
\begin{figure*}
    \centering
    \includegraphics[height=0.65\linewidth]{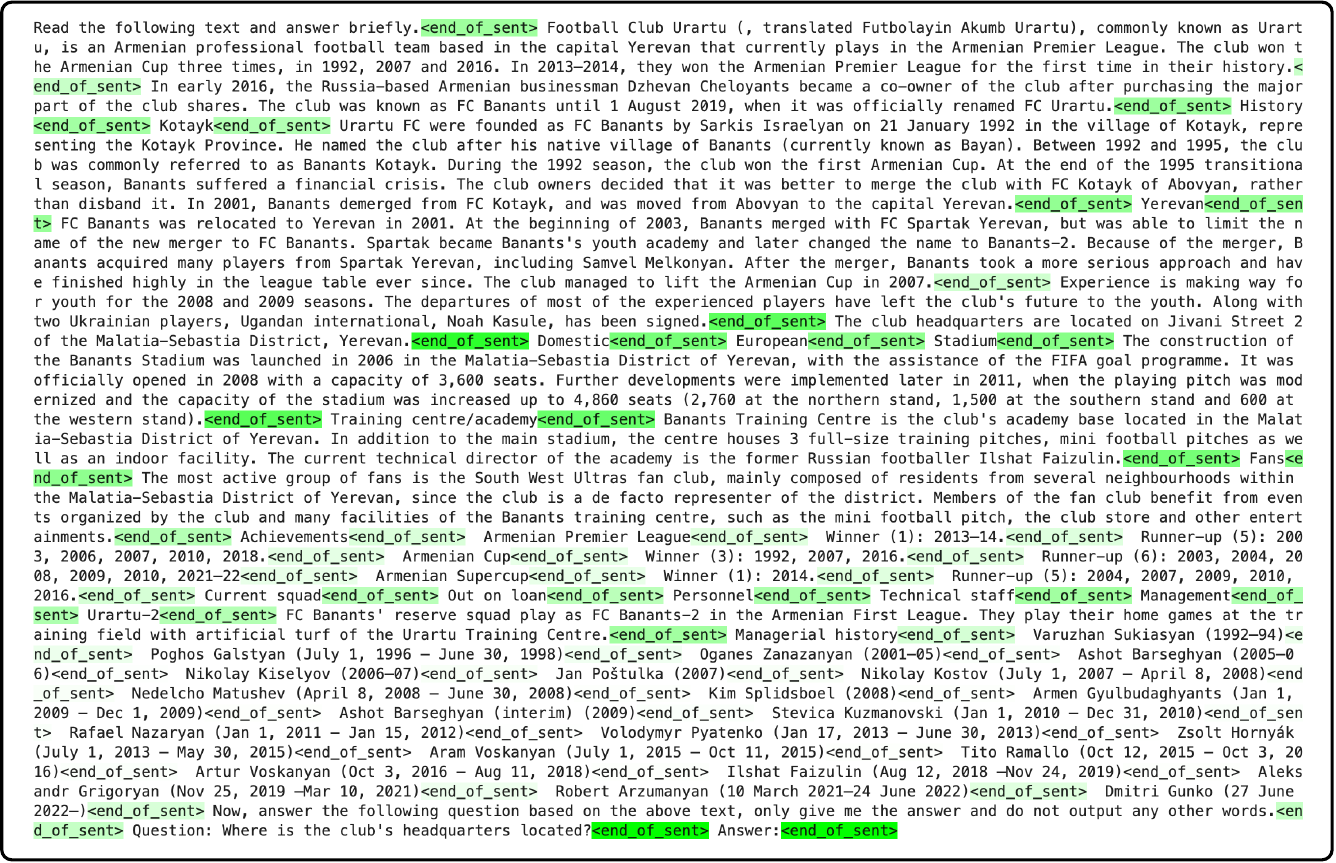}
    \caption{Visualization of the operation of the compression model with special tokens. Each \texttt{<end\_of\_sent>} token is highlighted with an intensity proportional to its importance in terms of answering the question.}
    \label{fig:spec_tokens}
\end{figure*}

Here, we provide additional qualitative examples. Figure \ref{fig:spec_tokens} shows the important of each sentence in the given context captured at the \texttt{<end\_of\_sent>} tokens. Here, darker colour indicates higher importance. 

\end{document}